\documentclass[conference]{IEEEtran}

\usepackage[letterpaper, left=0.625in, right=0.625in, bottom=1in, top=0.75in]{geometry}
\usepackage[accsupp]{axessibility} 
\usepackage{cite}
\usepackage{amsmath,amssymb,amsfonts}
\usepackage{algorithmic}
\usepackage{graphicx}
\usepackage{textcomp}
\usepackage{xcolor}
\usepackage{stfloats}
\usepackage{subcaption}
\usepackage{url}

\begin{document}


\title{A Quantitative Analysis of Multimodal\\Biomarkers in Alzheimer’s Disease}


\author {
\IEEEauthorblockN{Antonio Scardace}
\IEEEauthorblockA{\textit{Department of Mathematics and Computer Science} \\
\textit{University of Catania} \\ Catania, Italy \\ antonio.scardace@phd.unict.it}

\and

\IEEEauthorblockN{Daniele Ravì}
\IEEEauthorblockA{\textit{Department MIFT} \\
\textit{University of Messina} \\ Messina, Italy \\ daniele.ravi@unime.it}
}


\maketitle


\begin{abstract}
Despite increasing adoption of multimodal approaches in Alzheimer’s Disease (AD) research---aimed at integrating molecular, structural, clinical, and genetic biomarkers to enhance disease characterization---the relationships among these modalities remain poorly understood. A systematic analysis of their dynamic interaction is essential for improving disease modeling, identifying redundant assessments, and reducing patient burden and acquisition costs. In this paper, we present a quantitative analysis of multimodal AD biomarkers by integrating tau-PET, structural MRI, cognitive scores (MMSE and CDR), and APOE $\epsilon4$ data from 789 subjects drawn from the ADNI dataset. In our analyses, we (A) quantify cross‑modal mutual information and explained variance to assess redundancy and predictive dependencies; (B) examine associations between tau topologies and structural atrophy across brain regions to select informative ROIs; (C) perform a statistical decomposition of the tau–cognition association into atrophy‑related and atrophy‑independent components; (D) and identify a dominant neurodegenerative trajectory that aligns with cognitive decline. This study provides a systematic characterization of cross‑modal relationships, improving the interpretability and selection of biomarkers in AD. Code is publicly available at: \url{https://github.com/antonioscardace/Multimodal-AD}.
\end{abstract}




\section{Introduction}

Alzheimer’s Disease (AD) affects over 57 million people worldwide---a number projected to nearly triple by 2050~\cite{ADstats}. AD is characterized by a pathological cascade in which amyloid-$\beta$ accumulation is thought to trigger tau aggregation, which in turn leads to progressive neurodegeneration and brain atrophy, ultimately resulting in cognitive decline. Reflecting this biological progression, modern AD research increasingly relies on multimodal biomarkers to capture complementary aspects of disease pathology.

In both clinical practice and research, Magnetic Resonance Imaging (MRI) continues to play a central role in the evaluation of AD, allowing the assessment of macrostructural markers such as hippocampal volume loss and ventricular expansion~\cite{frisoni2010clinical}. These structural changes, however, typically reflect relatively advanced stages of the neurodegenerative process. Molecular imaging techniques, and in particular tau Positron Emission Tomography (tau-PET) with tracers such as $^{18}$F-AV-1451, instead provide a more direct assessment of AD pathology by detecting the accumulation of hyperphosphorylated tau. Tau deposition follows well-characterized neuroanatomical progression patterns and shows strong associations with subsequent brain atrophy and cognitive deterioration~\cite{mattsson201918f}. Complementing imaging biomarkers, clinical measures such as the Clinical Dementia Rating (CDR) and the Mini-Mental State Examination (MMSE) reflect the cognitive and functional consequences of the disease, capturing aspects that may not be fully explained by neuroimaging alone. In addition, genetic factors—most prominently the APOE $\epsilon4$ allele—represent the major genetic risk determinant for AD and substantially influence individual vulnerability to neurodegeneration~\cite{liu2013apolipoprotein}. The emergence of large multimodal datasets such as the Alzheimer’s Disease Neuroimaging Initiative (ADNI)~\cite{ADNI} has further propelled this field by enabling comprehensive analyses across well-phenotyped cohorts.

While the integration of these multimodal data has shown immense promise for improving diagnosis and prognosis~\cite{tauSynthesis2023,AnatomyGuided2025}, it often overlooks the fundamental question of the relative informational contribution of each modality, and whether redundant clinical assessments could be safely omitted. Addressing this gap is particularly important given the high cost of PET imaging, patient burden, and lengthy acquisition protocols, which motivate the identification of minimal yet maximally informative biomarker sets.

Most existing multimodal studies primarily focus on improving predictive accuracy through increasingly complex black-box models~\cite{Zhang2011,HoloDx}, often at the expense of interpretability and a quantitative understanding of cross-modal relationships. Recent efforts have begun to address this limitation by modeling biomarker interactions~\cite{Jin2016,Pontecorvo2017}. However, such approaches do not quantify causal pathways or the structure–pathology correspondence underlying disease progression, and generally overlook cross-modal relationships.

To address these limitations, we perform a systematic quantitative analysis of multimodal AD biomarkers, focusing on integrative interpretation of existing data modalities rather than the development of new methodological frameworks. Our goal is to analyze cross-modal relationships to improve interpretability and biomarker selection.

In particular, our main contributions are: (A.1) we quantify the cross‑modal mutual information to measure redundancy and isolate the unique contribution of each biomarker class, guiding biomarker selection; (A.2) we compute the cross‑modal explained variance to assess how well each modality can be predicted by the others; (B.1) we evaluate the intra‑regional correlations between localized tau uptake and structural volume, mapping the spatial footprint of tau toxicity and linking molecular pathology to local structural damage; (B.2) we examine the association between global topological patterns of tau accumulation and downstream structural atrophy across brain regions to guide the selection of ROIs with predictive relevance; (C) we perform a statistical decomposition of the tau–cognition association into atrophy‑related and atrophy‑independent components, estimating the proportion of the observed association that is attributable to atrophy‑related pathways; (D) we leverage the SuStAIn~\cite{SuStAIn,pySuStAIn} framework to identify the main neurodegenerative trajectory supported by the data, characterizing a dominant sequence of multimodal biomarker abnormalities which may help in staging and stratifying individuals within the cohort.


\section{Dataset Construction}

\begin{figure*}[!t]
    \centering
    \includegraphics[width=0.85\textwidth]{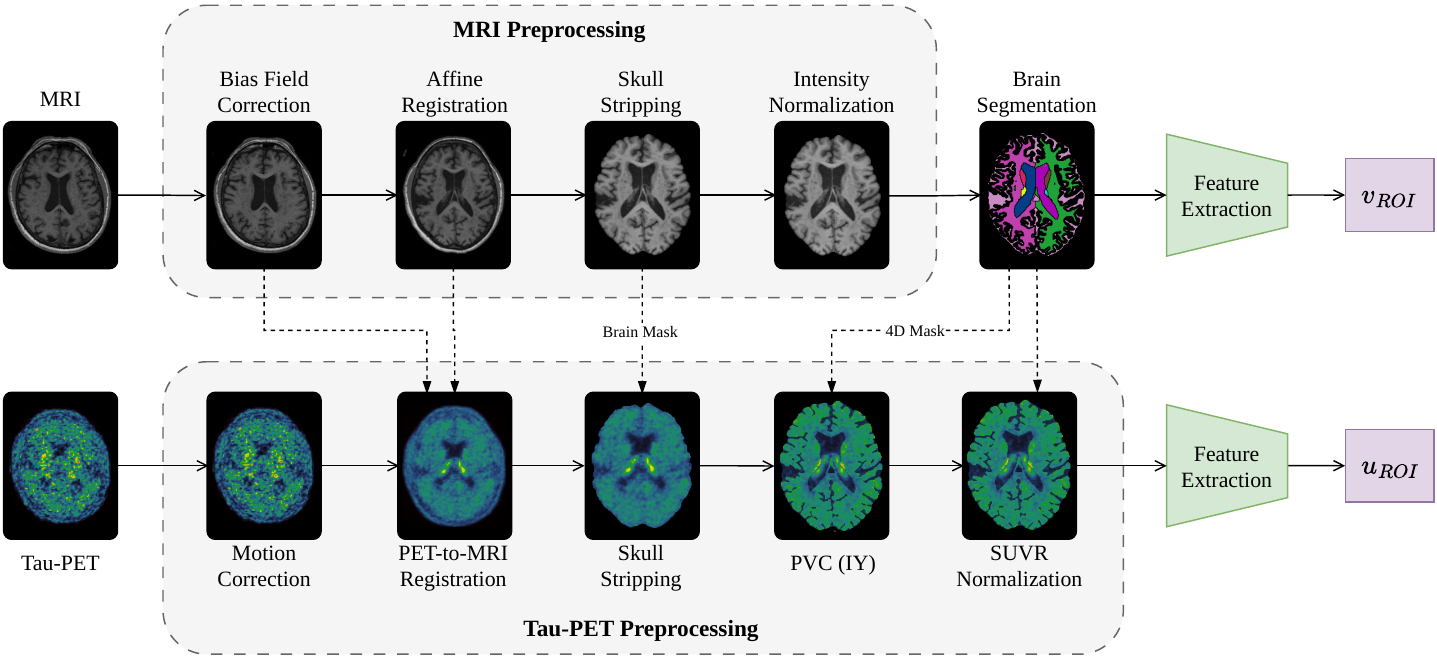}
    \caption{\label{fig:preproc} Overview of the multimodal preprocessing pipeline that operates through parallel branches to extract structural volumes ($v_{ROI}$) and molecular tau uptake ($u_{ROI}$).}
\end{figure*}

Our cross-sectional dataset includes 789 subjects from the ADNI 1/2/3/4/GO cohorts~\cite{ADNI}. Each visit has a T1w MRI, a tau-PET scan ($^{18}$F-AV-1451 radiotracer), the APOE $\epsilon4$ genotype, the MMSE and CDR scores. MRI and tau-PET scans are considered part of the same visit if acquired within a 6-month window~\cite{Scans6Months}, while MMSE and CDR scores are required within 12 months~\cite{Cdr12Months}. The cohort has a mean age of $74.1 \pm 8.3$ years, comprises 56.5\% females and 43.5\% males, and includes 59.9\% Cognitively Normal (CN), 34.7\% Mild Cognitive Impairment (MCI), and 5.3\% AD subjects.


\subsection{Preprocessing Pipeline}

In this section, we present our preprocessing pipeline, composed of two main blocks depicted in Figure~\ref{fig:preproc}.

MRI scans are preprocessed with bias field correction (ANTs~\cite{Ants2021}), affine registration (ANTs) to the MNI152 space, skull stripping (HD-BET~\cite{HdBet2019}), intensity normalization (WhiteStripe~\cite{WhiteStrip2016}), and segmentation of cortical and subcortical regions (SynthSeg~\cite{SynthSeg2023}) using the Desikan–Killiany atlas~\cite{DesikanKilliany}. All ROI volumes are normalized by total intracranial volume to ensure statistical rigor, and only the ROIs relevant to AD progression are retained. 

Tau-PET scans undergo motion correction and mean image computation for 4D acquisitions (ANTs), followed by PET-to-MRI registration and spatial normalization (ANTs), skull stripping, partial volume correction (PETPVC~\cite{PetPvc2016}) using the Iterative-Young algorithm to correct spill-over between regions, and Standardized Uptake Value Ratio (SUVR) normalization using the cerebellum as reference. We retained only visits with mean tau uptake within the physiologically plausible range of $[0,3]$ SUVR to exclude extreme outliers~\cite{PetRange13}. Finally, mean uptake is selected instead of total uptake to robustly decouple the molecular measurement of tau pathology from the confounding effects of structural atrophy.


\section{Analysis}

This section reports the findings from our in‑depth analysis of the multimodal dataset considered in this study. In particular, we first quantify cross‑modal redundancy and predictive complementarity in Section~\ref{sec:info}, next characterize regional associations between tau accumulation and structural atrophy to identify informative ROIs in Section~\ref{sec:net}, then use statistical decomposition to link tau pathology and atrophy to cognitive decline in Section~\ref{sec:causal}, and finally apply trajectory modeling to reconstruct the dominant disease progression pattern in Section~\ref{sec:sustain}.


\subsection{Quantifying Cross-Modal Information Sharing} \label{sec:info}

We quantify how tau-PET, MRI, APOE $\epsilon4$, MMSE, CDR-SB, and CDR-GLOBAL modalities share information and predict each other. Specifically, we use two complementary measures: Normalized Mutual Information (NMI) and cross-modal explained variance ($R^2$).

\vspace{0.4cm}
\subsubsection{Quantifying Cross-Modal Redundancy}

\begin{figure}[h!]
    \centering
    \includegraphics[width=\columnwidth]{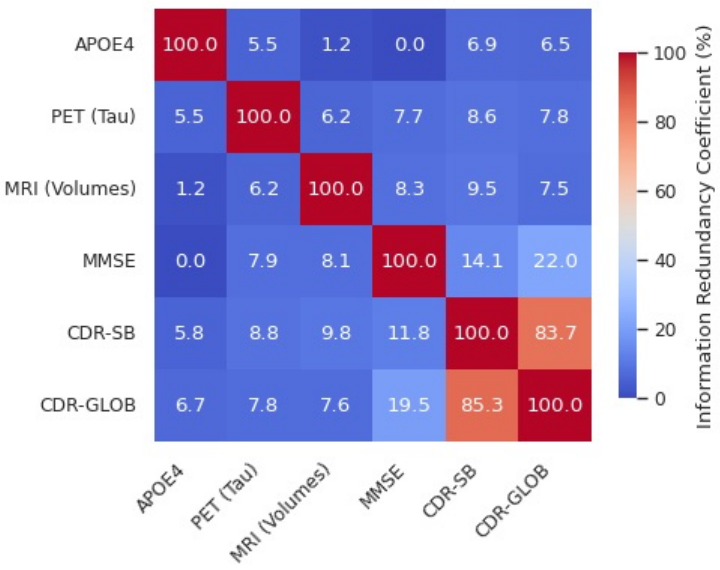}
    \caption{\textbf{Normalized Mutual Information across Biological and Clinical Domains.} Cells in the matrix represent the Information Redundancy Coefficient (\%). Higher values indicate greater shared information. The matrix reveals that clinical assessments are highly redundant, whereas the genetic profile (APOE $\epsilon4$) and neuroimaging modalities contribute predominantly orthogonal diagnostic information.}
    \label{fig:redundancy}
\end{figure}

To assess statistical dependence without assuming linear relationships, we compute NMI between each pair of modalities using a non-parametric nearest-neighbor estimator~\cite{Kraskov2004}. Mutual information is normalized by the minimum marginal entropy~\cite{Estevez2009} to obtain a redundancy coefficient bounded between 0\% and 100\%, facilitating direct interpretability across modality pairs. As shown in Figure~\ref{fig:redundancy}, APOE $\epsilon4$ shares negligible information with all the other modalities (0.0--6.9\%), confirming its largely independent and complementary contribution. Similarly, MRI and tau-PET exhibit low shared information (6.2\%) and limited overlap with cognition (7.5--9.8\%). In stark contrast, clinical scores are highly redundant with each other. In fact, CDR-SB and CDR-GLOBAL share more than 83\% of their information, confirming their substantial overlap.

\subsubsection{Quantifying Cross-Modal Explained Variance}

\begin{figure}[h!]
    \centering
    \includegraphics[width=\columnwidth]{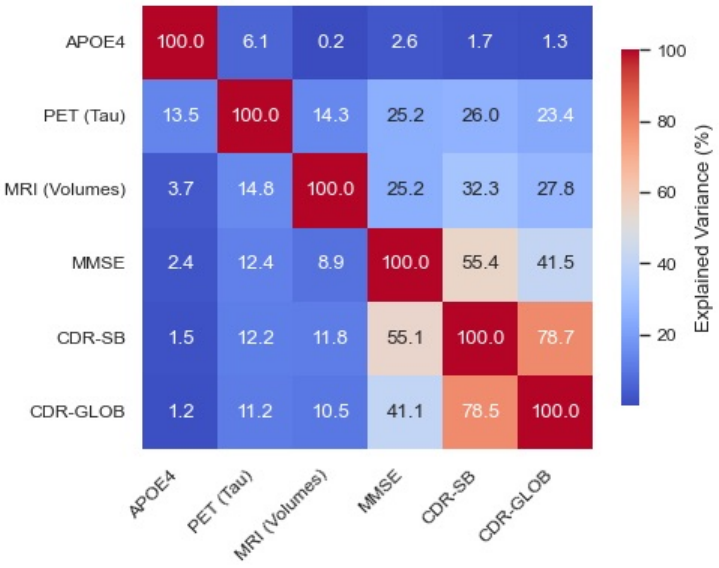}
    \caption{\textbf{Cross-Modal Directional Explained Variance.} Cells in the matrix represent the percentage of variance ($R^2$) in a target modality (columns) explained by a predictor modality (rows) using cross-validated Ridge regression. This asymmetry demonstrates a clear directional relationship: structural and molecular changes are strong predictors of clinical outcomes, whereas clinical assessments reflect aggregate functional measures that cannot recover spatial patterns of neurodegeneration.}
    \label{fig:variance}
\end{figure}

To evaluate directional relationships, we estimate cross-modal explained variance using cross-validated Ridge regression, with the regularization parameter selected independently for each predictor–target pair through internal cross-validation over 100 values uniformly sampled in log-space in the interval $[10^{-4}, 10^{4}]$. Each modality is predicted from the others, producing an $R^2$ matrix (Figure~\ref{fig:variance}). All features in this analysis were standardized using z-score normalization to ensure comparability across modalities. In our results, we can see that APOE $\epsilon4$ acts as an upstream variable: it weakly predicts tau burden (6.1\%) but cannot be inferred from downstream structural or cognitive measures. Neuroimaging biomarkers instead show substantial predictive power over cognition. MRI explains 32.3\% of CDR-SB variance, while tau-PET explains 25.2\% of MMSE variance. Reverse mappings are markedly weaker; in particular, molecular and structural alterations constrain cognitive decline, whereas clinical scores represent aggregate outcomes that cannot uniquely recover the spatial distribution of the underlying pathology. We interpret these results as directional statistical associations rather than causal effects, with the limitation that the model may not fully capture nonlinear dependencies. The choice of Ridge regression with cross-validation was motivated by its ability to mitigate overfitting in the presence of substantial multicollinearity, as many biomarkers are strongly correlated and tend to change jointly during neurodegeneration.

\subsection{Structure--Pathology Associations Across Brain Regions}~\label{sec:net}

\subsubsection{Intra-Regional}

\begin{figure}[h!]
    \centering
    \includegraphics[width=\columnwidth]{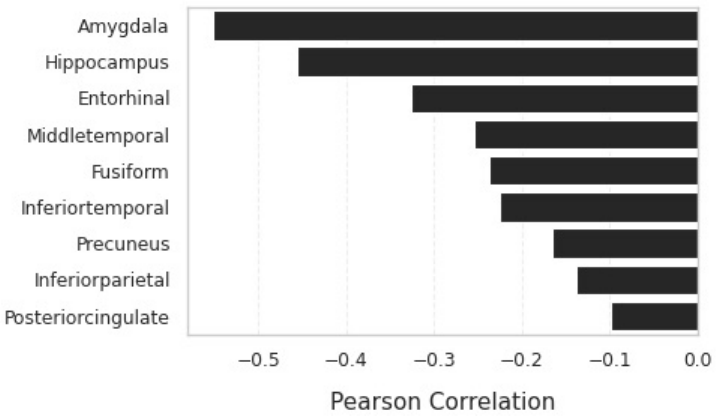}
    \caption{\textbf{Intra-Regional Structure--Pathology Associations.} The bar chart displays the Pearson correlation coefficients ($r$) between mean tau uptake and structural volume for each cortical and subcortical ROI on the horizontal axis, ranked by effect magnitude. All associations are statistically significant following FDR correction.}
    \label{fig:correlation}
\end{figure}

To investigate regional associations between molecular tau accumulation and atrophy, we computed the intra-regional Pearson correlation between mean tau uptake and MRI volume for each ROI. To account for multiple comparisons, $p$-values were adjusted using the False Discovery Rate (FDR) procedure~\cite{Fdr1995}. As illustrated in Figure~\ref{fig:correlation}, all examined regions show a statistically significant negative correlation (FDR-adjusted $p < 0.05$), confirming that regions with higher tau uptake tend to show greater volume reduction.

\subsubsection{Cross-Regional}

\begin{figure}[!t]
    \centering
    \includegraphics[width=\columnwidth]{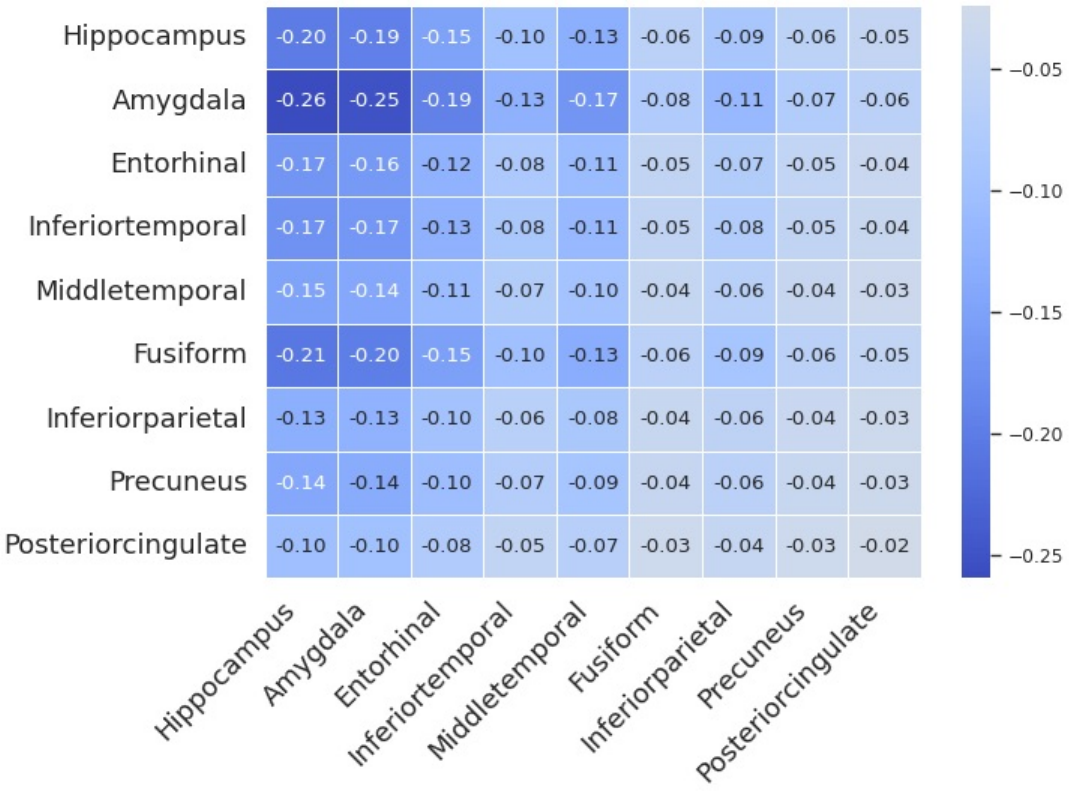}
    \caption{\textbf{Cross-Regional Tau–Atrophy Associations.} Cells in the matrix quantify how strongly tau accumulation in a specific region (rows) is coupled with structural volume loss in another region (columns) on average across the entire cohort. Only interaction weights with 95\% CI (derived from 7,500 bootstrap iterations) are retained.}
    \label{fig:network}
\end{figure}

Extending beyond local effects, we used Partial Least Squares Singular Value Decomposition (PLS-SVD)~\cite{PlsSvd2011} to capture multivariate tau–atrophy associations across brain regions. After regressing out age and sex, latent components maximizing the covariance between residual tau-PET uptake and MRI volumes were extracted. Robustness was assessed via non-parametric bootstrap resampling (7,500 iterations), retaining only weights with 95\% confidence intervals (CI). The resulting cross-modal interaction matrix (Figure~\ref{fig:network}) shows negative associations between tau accumulation in any brain region and volume loss in all other regions. This reveals a global pattern where higher tau accumulation correlates with widespread brain atrophy across regions.

\subsection{From Tau to Cognition: The Role of Atrophy}~\label{sec:causal}

To elucidate the association linking molecular pathology to cognitive decline, we performed a statistical decomposition analysis~\cite{Imai2010} to assess whether structural atrophy accounts for part of the tau–cognition association. Global tau burden was defined as the independent variable, the global atrophy index as the mediator, and CDR-SB as the clinical outcome, adjusting for age and sex. Confidence intervals were estimated via non-parametric bootstrap resampling (7,500 iterations). Biomarker indices for tau and atrophy were z-standardized prior to analysis. Tau pathology showed a significant total effect on cognition: a one-standard-deviation increase in global tau was associated with a 1.006-point increase in CDR-SB.

In our experiments, we use the Average Causal Mediation Effect (ACME) to quantify the indirect pathway through structural atrophy and to assess how much of tau’s association with cognition operates via macroscopic neurodegeneration. The ACME was significant ($0.284$, 95\% CI: $[0.197, 0.384]$), suggesting that structural neurodegeneration mediates approximately 28\% of tau’s total association with cognition. However, given the cross-sectional design, these findings should be interpreted cautiously, as they do not support strong causal claims and remains sensitive to unmeasured confounding.

The Average Direct Effect (ADE) captures the remaining direct association between tau pathology and cognition beyond the contribution explained by volume loss, and may reflect additional mechanisms such as functional network disruption. The ADE was also significant ($0.722$, 95\% CI: $[0.546, 0.907]$), confirming a substantial direct tau–cognition association beyond macroscopic atrophy.


\subsection{Data-Driven Modeling of Disease Progression}~\label{sec:sustain}

While cross-regional tau–atrophy associations (see Section~\ref{sec:net}) characterize population-level relationships, they provide a static view and do not capture temporal event progression. To address this limitation, we applied SuStAIn~\cite{SuStAIn}, an unsupervised algorithm that reconstructs disease trajectories from cross-sectional data.

SuStAIn models disease progression as a sequence of discrete abnormality events; therefore, continuous biomarker measurements must first be mapped onto a common severity scale. For this reason, we applied normative modeling with demographic deconfounding via linear regression. After regressing out age, sex, and total intracranial volume, biomarker residuals were $z$-standardized relative to the CN reference group. This expresses each biomarker as a standardized deviation from the CN distribution, where $z=0$ corresponds to the CN mean and each unit represents one standard deviation of the reference population. To explicitly model the known biological lag between molecular pathology and downstream neurodegeneration, disease progression was parameterized using modality-specific abnormality thresholds: $z \in \{1, 2, 3\}$ for tau-PET~\cite{SuStAIn} and delayed thresholds of $z \in \{2, 3, 4\}$ for structural MRI. The use of $z=2$ as the first structural abnormality threshold follows the normative modeling framework~\cite{Marquand2016}, where values that differ from the healthy population mean by more than two standard deviations (approximately the 95th percentile) are commonly interpreted as pathological outliers rather than normal variability. This choice increases specificity by separating normal aging-related brain volume variability from more pronounced atrophy likely related to disease. Each SuStAIn stage corresponds to the acquisition of one additional biomarker abnormality event. Given 9 ROIs, 2 modalities (tau-PET and MRI), and 3 severity levels per biomarker (mild, moderate, severe), the model defines a total of 54 abnormality events ($9\times2\times3$), resulting in 54 stages.

\begin{figure*}[!t]
    \centering
    \includegraphics[width=\textwidth]{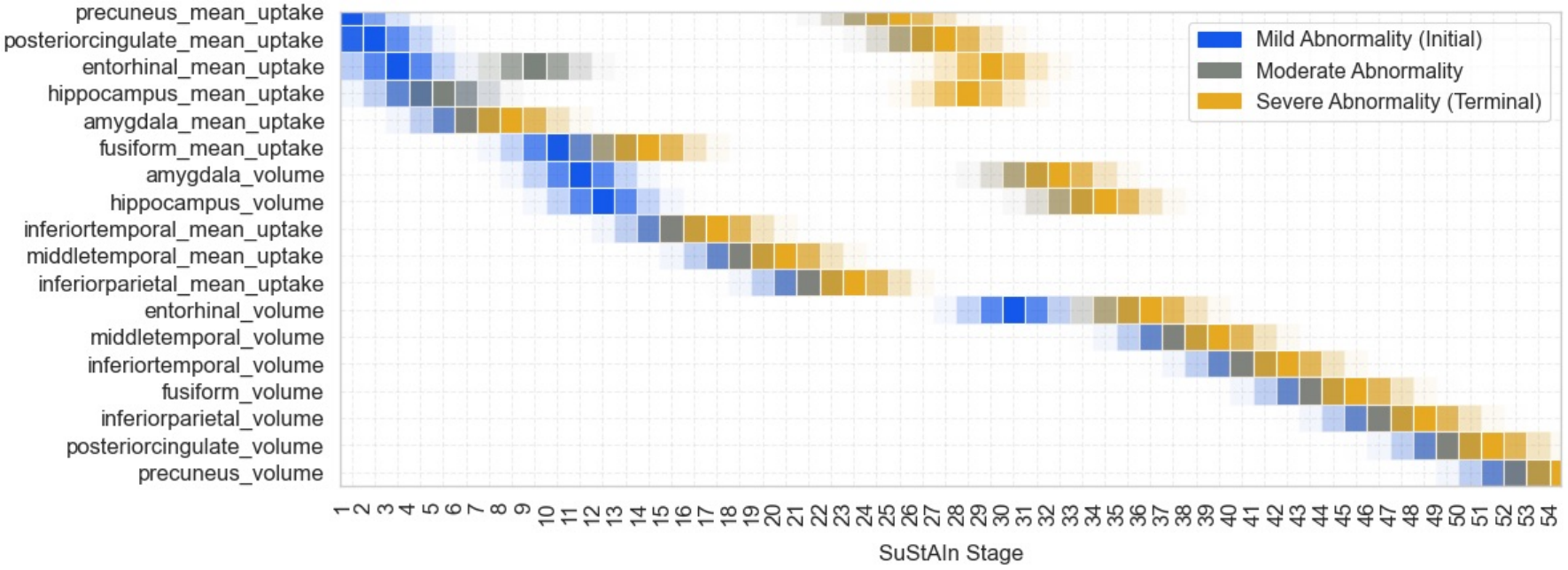}
    \caption{\textbf{Temporal Cascade of Biomarker Abnormalities.} The vertical axis lists the multimodal biomarkers, while the horizontal axis represents the SuStAIn stages. Colors indicate the transition to mild (blue), moderate (gray), and severe (yellow) abnormality thresholds. Block opacity reflects the MCMC sampling probability of that specific event occurring at a given stage: solid, vibrant colors denote high certainty in the temporal ordering, whereas faded, semi-transparent blocks indicate temporal variance (uncertainty) across the cohort. Empty cells signify a near-zero probability.}   
    \label{fig:stages}
\end{figure*}

The spatiotemporal architecture of the dominant trajectory (Figure~\ref{fig:stages}) demonstrates how the model reconstructs plausible pseudo-temporal progressions from heterogeneous data. The positional variance diagram suggests a plausible pseudo-temporal cascade in which the inferred ordering is consistent with tau abnormalities emerging before detectable structural atrophy in medial temporal regions detected in MRI scans.

\section{Discussion and Conclusion}

In this work, we presented a systematic quantitative analysis to analyze the multimodal biomarker landscape of AD. Our objective was to provide an interpretable and quantitative mapping of how genetic, molecular, structural, and clinical modalities interact, share information, and jointly reflect the neurodegenerative cascade. This type of analysis is particularly relevant to avoid black-box multimodal models for AD that are often designed without a clear understanding of cross-modal redundancy and the relative contribution of each modality.

We observed that global clinical assessments such as CDR and MMSE share substantial informational overlap, suggesting that administering multiple global cognitive scales may introduce unnecessary costs without proportional diagnostic benefit. Conversely, genetic profiling (APOE $\epsilon4$) and neuroimaging modalities provide largely complementary information, supporting their joint use in multimodal modeling. Furthermore, our statistical decomposition analysis provides evidence of an atrophy-related indirect component in the tau–cognition association, but a substantial direct component remains, suggesting that the tau–cognition association may also involve additional mechanisms independent of macroscopic volume loss.

We further showed that the observed associations between tau accumulation and structural atrophy may reflect a temporally delayed cascade, highlighting the importance of modeling multimodal biomarkers as temporally offset biological processes rather than as synchronous features.

Despite these insights, this study presents some limitations, including the reliance on cross-sectional data, the very limited number of AD subjects, and the exclusive use of the ADNI cohort, which may limit generalizability to real-world and more heterogeneous clinical populations. Another limitation is the exclusion of amyloid‑$\beta$ and other modalities, which, although biologically central to the AD cascade, would have drastically reduced the available cohort size. Future work should address these constraints by integrating longitudinal data to better capture temporal dynamics, incorporating amyloid‑$\beta$ and other key modalities, and validating the proposed framework on independent external cohorts to assess its robustness and generalizability across diverse clinical settings.

In conclusion, by integrating multimodal biomarkers and quantifying their statistical dependencies, this work provides a transparent and biologically informed framework for designing multimodal AI systems in AD. Our analyses highlight population‑level patterns of neurodegeneration, the temporal relationships between molecular and structural changes, and the importance of aligning computational models with the underlying biological cascade. From a machine learning perspective, these findings suggest that modality selection should prioritize complementary biological sources and informative brain regions rather than maximizing the number of inputs.


\section*{Acknowledgements}

This work was partially funded by the Italian Ministry of University and Research (MUR) under the Fondo Italiano per la Scienza (FIS 3), project “AI-Powered Digital Twins for Neurodegeneration Research: Advancing Disease Progression Modeling (NeuroTwin)” (CUP J53C25002200001).


\bibliographystyle{IEEEtran}
\bibliography{IEEEabrv, IEEEbibliography}


\end{document}